\documentclass{article}
\usepackage[preprint]{neurips_2026}
\usepackage{graphicx} % Required for inserting images
\usepackage{multirow}
\usepackage{booktabs}
\usepackage{subcaption}
\usepackage{url}
\usepackage[hidelinks]{hyperref}
\usepackage{amsmath,amssymb}
\usepackage{xcolor}

\hypersetup{hidelinks}

\definecolor{algblue}{HTML}{1A73E8}

\usepackage{algorithm}
\usepackage[noend]{algpseudocode}

\newcommand{\codeurl}{\url{https://github.com/joesharratt1229/ThriftAttention}}

\algrenewcommand{\algorithmiccomment}[1]{\hfill// #1}
\title{ThriftAttention: Selective Mixed Precision for Long-Context FP4 Attention}
\author{
Joe Sharratt \\
\texttt{joesharratt29@gmail.com} \\
\codeurl
}

\begin{document}
\maketitle
\begin{abstract}
Efficient attention algorithms are critical to mitigate the quadratic cost of attention in long-context workloads. Prior work utilises block-scaled quantisation techniques on Blackwell GPUs to move attention computation to 4-bit precision to accelerate inference. However, these techniques result in significant quality degradation in long-context settings. We show that the output impact of quantisation error is highly non-uniform and increases with the importance of each query-key interaction, concentrating functionally relevant error in a small number of attention blocks that contain the most important tokens. We propose \textbf{ThriftAttention}, a low-bit attention variant that delivers near-FP16 long-context quality at FP4 inference efficiency. This approach proceeds in two stages. \textbf{(1)} A heuristic rapidly selects a small number of important query-key block pairs for FP16 precision. \textbf{(2)} The selected blocks are computed in FP16 and the remaining blocks in FP4, with both paths merged via online softmax into a single output. We demonstrate across long-context benchmarks and model families that by computing only $5\%$ of query-key blocks in FP16, ThriftAttention recovers on average $89.1\%$ of the FP4$\to$FP16 performance gap. We show ThriftAttention's advantage grows with sequence length, mitigating the systematic FP4 quality degradation observed at longer contexts. The code is available at \url{https://github.com/joesharratt1229/ThriftAttention}.
\end{abstract}
                   
\section{Introduction}
Efficient inference is critical for the deployment of large language models \citep{pope2023efficiently, wan2024efficient}. Attention \citep{vaswani2017attention, zhang2025efficient_attention_survey} is a key bottleneck in long-context workloads, where its quadratic cost and KV-cache memory traffic dominate execution time \citep{kwon2023efficient, patel2024splitwise}. NVIDIA's Blackwell architecture \citep{nvidia2024blackwell} introduces native FP4 Tensor Cores that have $4\times$ the arithmetic throughput of the equivalent FP16 instructions on Blackwell GPUs \citep{nvidia_ptx_isa_9_2} whilst reducing KV-cache memory traffic by a similar amount. Recent works, including SageAttention3 \citep{zhang2025sageattention3}, exploit this hardware to accelerate attention. This introduces a fundamental tension between inference efficiency and output quality.

\begin{figure*}[t]
\small
  \centering
  \includegraphics[width=0.6\textwidth]{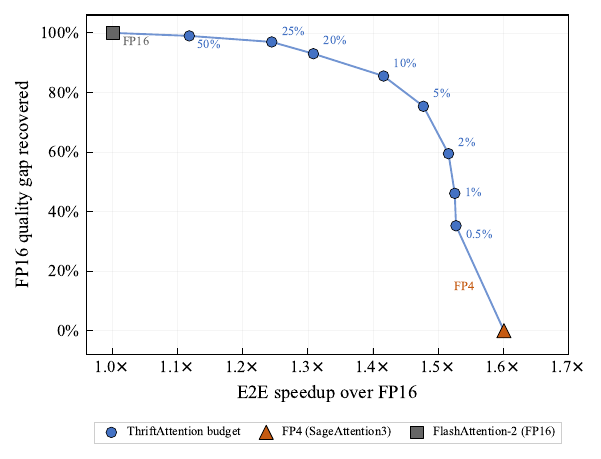}
  \caption{
  \textbf{ThriftAttention approaches FP4 latency while preserving near-FP16 quality.}
 Pareto frontier of negative log-likelihood (NLL) recovery vs inference efficiency at 131k context length (Qwen3-8B). Performance recovery is measured as the percentage of the FP4-to-FP16 NLL gap recovered.
  }
  \label{fig:pareto-e2e-decode}
\end{figure*}                                                               
Two lines of prior work address parts of this problem but neither fully resolves it. FP4 attention methods accept the quality degradation as a cost of increased throughput. Sparsity methods \citep{tang2024quest, zhang2025spargeattn, zhang2023h2o} take the approach of identifying important query-key interactions and computing only those interactions. However, sparsity approaches must drop at least $75\%$ of KV blocks to match FP4 latency during the generation phase at inference. Such aggressive sparsity ratios can become a major source of performance degradation in inference-only sparsity methods because the error from omitting blocks entirely is irrecoverable.

In this work we develop \textbf{ThriftAttention}, a training-free mixed-precision attention mechanism that delivers near-FP16 long-context quality at FP4 inference efficiency. This addresses the degradation of low-bit attention variants at long contexts. We show that functionally relevant quantisation error is not evenly distributed across query-key interactions. Instead, it is concentrated in a small number of interactions where the attention scores are high in magnitude and most important to the final output distribution. 

This finding motivates a simple two-stage approach. First, a lightweight heuristic scores each query-key block pair by $\hat{S}_{ij} = \bar{q}_i \cdot \bar{k}_j$, where $\bar{q}_i$ and $\bar{k}_j$ are the token means of each query block $q_i$ and key block $k_j$ respectively. The top-$k$ highest-scoring blocks are selected for FP16 precision whilst the remaining query-key block computations are assigned to FP4. Attention is computed for both sets of blocks and then combined via online softmax into a single output. 

\paragraph{Results.} ThriftAttention recovers most of the quality gap between FP4 and FP16 attention whilst retaining the efficiency benefits of FP4 inference.  At a $5\%$ FP16 block budget, ThriftAttention recovers on average $89.1\%$ of the FP4$\to$FP16 performance gap. This increases to $91.8\%$ and $92.4\%$ recovery at $10\%$ and $25\%$ respectively. In end-to-end generation, ThriftAttention reduces inference latency by up to $2\times$ at long contexts. Sequence-length analysis demonstrates ThriftAttention's advantage grows with context length, where uniform FP4 attention degrades most severely.

\paragraph{Contributions.} Our work makes the following contributions:
\begin{itemize}
\item We introduce ThriftAttention, a training-free attention approach that computes the most important block interactions in FP16 and the remainder in FP4. To our knowledge, this is the first work to use sub-byte formats in this mixed-precision manner for attention computation.
\item We evaluate ThriftAttention on LongBench-v1, HELMET, RULER, and PG-19 across Llama, Qwen, and Ministral model families, showing that a small FP16 budget recovers most of the FP16 quality while retaining low-bit inference efficiency.
\end{itemize} 

\begin{figure*}[t]
\small
  \centering
  \includegraphics[width=\textwidth]{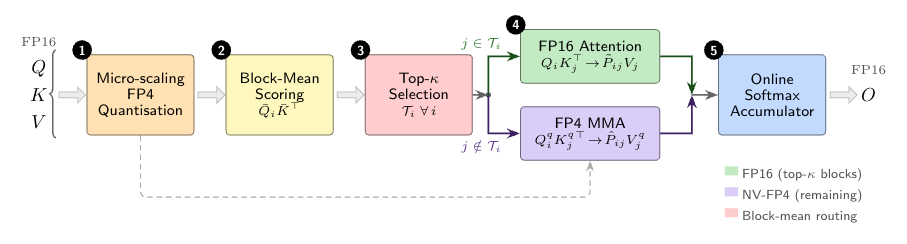}
  \caption{Overview of \textbf{ThriftAttention}}
  \label{fig:architectural-diagram}
\end{figure*}

\section{Related Work}
\textbf{I/O-Efficient Attention.} FlashAttention~\citep{dao2022flashattention} introduced tiling to reduce GPU memory I/O, with subsequent versions~\citep{dao2024flashattention2,shah2024flashattention3,zadouri2026flashattention4} improving parallelism and adding hardware-specific optimisations.

\textbf{Quantised Attention.}
While post-training quantisation is well-established for linear
layers~\citep{dettmers2022llmint8, frantar2022gptq, lin2023awq,
dettmers2023qlora, ashkboos2024quarot, liu2024spinquant}, its extension to
attention remains limited. SageAttention~\citep{zhang2025sageattention,
zhang2025sageattention2} accelerates attention via INT8/FP8 quantisation
with outlier smoothing, and SageAttention3~\citep{zhang2025sageattention3}
extends this to FP4 on Blackwell using two-level microscaling.
Other works target KV cache compression~\citep{liu2024kivi,
hooper2024kvquant, lin2024qserve} or combine quantised matmuls with
sparsity~\citep{kang2024turboattention}. 

\textbf{Sparse Attention.} Quest~\citep{tang2024quest} performs query-aware KV block selection using coordinate-wise min-max bounds. Token-level
eviction and selection strategies~\citep{zhang2023h2o,
xiao2024streamingllm, li2024snapkv, jiang2024minference} reduce the
active KV set, while NSA~\citep{yuan2025nsa} and
SLA~\citep{zhang2025sla} learn sparse structure during training.
SpargeAttn~\citep{zhang2025spargeattn} combines block-sparsity prediction
with quantised attention, skipping near-zero blocks before computing
the remainder in INT8/FP8. This is the closest prior work to ours but
uses sparsity as the primary acceleration mechanism and does not
utilise sub-8-bit numeric formats. Other approaches including linear
attention~\citep{katharopoulos2020transformers, choromanski2021rethinking,
wang2020linformer, qin2024lightning, yang2024gated} work less
effectively where the attention distribution is peaked.

\textbf{Positioning.} ThriftAttention allocates full precision to the most important blocks rather than imposing uniform quantisation. For a given block the error is upper-bounded by FP4 quantisation noise rather than the magnitude of skipped/approximated attention scores as in sparse methods.

\section{Method}
\begin{figure*}[t]
\small
  \centering
  \includegraphics[width=0.65\textwidth]{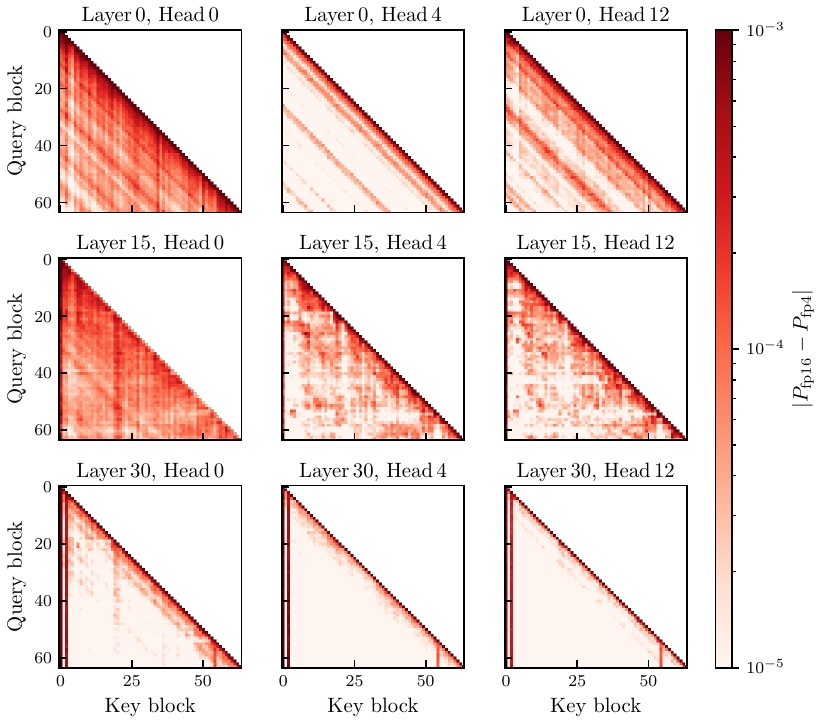}
  \caption{Typical FP16$\to$FP4 attention quantisation error, $e = \left|P_{\mathrm{FP16}} - P_{\mathrm{FP4}}\right|$, by query/key blocks across layers and heads in Qwen3-8B (seq=4096)}
  \label{fig:quantisation_error}
\end{figure*}
\subsection{Motivation for ThriftAttention}

Consider a single query token attending over $N$ keys. The attention output is
\begin{equation}
o = \sum_{j=1}^{N} p_j \, v_j, \qquad p_j = \frac{\exp(s_j)}{\sum_{k} \exp(s_k)}, \qquad s_j = q \cdot k_j / \sqrt{d}.
\end{equation}
FP4 quantisation perturbs each score by some $\epsilon_j$, giving $\tilde{s}_j = s_j + \epsilon_j$. The first-order output perturbation is
\begin{equation}
\label{eq:perturbation}
\delta o = \tilde{o} - o \approx \sum_{j} \frac{\partial o}{\partial s_j} \, \epsilon_j.
\end{equation}
From the softmax Jacobian ($\partial p_j / \partial s_j = p_j(1 - p_j)$, $\partial p_k / \partial s_j = -p_k p_j$ for $k \neq j$):
\begin{equation}
\label{eq:sensitivity}
\frac{\partial o}{\partial s_j} = p_j (v_j - o).
\end{equation}
Substituting and taking norms:
\begin{equation}
\label{eq:error_bound}
\|\delta o\| \leq \sum_{j} |\epsilon_j| \cdot p_j \cdot \|v_j - o\|.
\end{equation}
Each key's error contribution is the product of three terms: $|\epsilon_j|$, the score quantisation error; $p_j$, the attention weight; and $\|v_j - o\|$, the value deviation from the output. The $p_j$ factor makes this non-uniform such that tokens with large pre-softmax scores produce large $p_j$ through the softmax exponential, amplifying their own quantisation error. Conversely low attention scores dampen the effect of a key token's quantisation error on $\tilde{o}$.

Figure \ref{fig:quantisation_error} illustrates this structure. Visualising $e = |P_{\text{FP16}} - P_{\text{FP4}}|$ across query-key block pairs for representative layers and heads in Qwen3-8B, the error concentrates in a small number of blocks per query. These are typically near-diagonal blocks and non-initial attention sinks, precisely where attention scores are largest.

This concentration suggests that most of the quality loss from uniform FP4 attention can be recovered by selectively promoting only these high-error blocks to FP16 and keeping the rest in FP4. This is most critical at long contexts, where per-token error compounds across more positions.

\subsection{ThriftAttention Algorithm}
\paragraph{FP4 quantisation.}
Let $X \in \mathbb{R}^{N \times d}$. We quantise $X$ to an FP4 tensor
$X^{q} \in \mathbb{R}^{N \times d}$ together with an FP8 microscale tensor
$S_X \in \mathbb{R}^{N \times d/16}$. We use the NVFP4 microscaling format \citep{nvidia2024blackwell, rouhani2023microscaling}
supported on Blackwell GPUs, in which each element of $X^{q}$ is stored in
E2M1 format and each per-group scale in $S_X$ is stored in E4M3 format.
This quantisation is applied independently to $Q$, $K$, and $V$, where
$Q, K, V \in \mathbb{R}^{N \times d}$.

\paragraph{Block-importance scoring.}
We partition $Q$ into $T_q = N/B_q$ blocks and $K$, $V$ into
$T_k = N/B_k$ blocks:
\[
Q = [Q_1; \dots; Q_{T_q}], \qquad
K = [K_1; \dots; K_{T_k}], \qquad
V = [V_1; \dots; V_{T_k}],
\]
where $Q_i \in \mathbb{R}^{B_q \times d}$ and
$K_j, V_j \in \mathbb{R}^{B_k \times d}$.
Let $\mathcal{B}_i^Q$ and $\mathcal{B}_j^K$ denote the token index sets of
query block $i$ and key/value block $j$, respectively. We compute the
token-wise mean of each block:
\[
\bar{Q}_i = \frac{1}{B_q}\sum_{t \in \mathcal{B}_i^Q} Q_t,
\qquad
\bar{K}_j = \frac{1}{B_k}\sum_{t \in \mathcal{B}_j^K} K_t.
\]
The importance score for a block pair $(i,j)$ is then
\[
\hat{S}_{ij} = \bar{Q}_i \bar{K}_j^\top .
\]

\paragraph{Mixed-precision attention computation.}
For each query block $i$, we select the top-$k$ key blocks
$\mathcal{T}_i = \operatorname{TopK}(\{\hat{S}_{ij}\}_{j=1}^{T_k},\, k)$.
Query-key block pairs are routed to the FP4 or FP16 path:
\[
S_{ij} =
\begin{cases}
Q_i K_j^\top / \sqrt{d}, & j \in \mathcal{T}_i, \\[3pt]
\operatorname{Matmul}_{\mathrm{FP4}}(Q_i^{q}, K_j^{q}, S_{Q,i}, S_{K,j})/\sqrt{d},
& j \notin \mathcal{T}_i.
\end{cases}
\]
\[
\widetilde P_{ij} = \operatorname{OnlineSoftmax}(S_{ij}).
\]
The output accumulation follows two paths. For $j \in \mathcal{T}_i$, we follow the standard FlashAttention-2 online softmax \citep{dao2024flashattention2} procedure. For $j \notin \mathcal{T}_i$, the probability block is quantised via the two-level scheme of SageAttention3 \citep{zhang2025sageattention3}:
\[
(\widehat P_{ij},\, S^{(2)}_{P,ij}) = \phi(\widetilde P_{ij}/s^{(1)}_{P,ij}), \qquad
O_i \mathrel{+}= s^{(1)}_{P,ij} \cdot \operatorname{Matmul}_{\mathrm{FP4}}(\widehat P_{ij},\, V_j^q,\, S^{(2)}_{P,ij},\, S_{V,j}),
\]
where $s^{(1)}_{P,ij} = \operatorname{rowmax}(\widetilde P_{ij})/(448 \times 6)$. The FP16 and FP4 output updates are merged online. The full procedure is given
in Algorithm~\ref{alg:mixed}, which presents the non-causal version. For causal LLM
attention, we apply the standard causal mask and restrict block selection to causally visible key blocks.

\begin{algorithm}[t]
\caption{Forward pass of ThriftAttention.}
\label{alg:mixed}
\begin{algorithmic}[1]
\Statex \textbf{Input:} $Q,K,V \in \mathbb{R}^{N \times d}$; block sizes $B_q,B_k$; FP16 budget $k$
\State Partition $Q$ into $\{Q_i\}_{i=1}^{T_q}$ and $K,V$ into $\{K_j\}_{j=1}^{T_k}, \{V_j\}_{j=1}^{T_k}$
\State $(Q^{q},S_Q)\gets \operatorname{fp4}(Q)$;\; $(K^{q},S_K)\gets \operatorname{fp4}(K)$;\; $(V^{q},S_V)\gets \operatorname{fp4}(V)$
\State Compute block means $\{\bar Q_i\}_{i=1}^{T_q}$ and $\{\bar K_j\}_{j=1}^{T_k}$
\For{$i=1$ to $T_q$}
    \State $\mathcal T_i \gets \operatorname{TopK}(\{\bar Q_i \bar K_j^\top\}_{j=1}^{T_k}, k)$
    \State $m_i^{(0)} \gets -\infty$;\; $\ell_i^{(0)} \gets \mathbf{0}$;\; $O_i \gets \mathbf{0}$
    \For{$j=1$ to $T_k$}
        \If{$j \in \mathcal T_i$}
            \State $S_{ij} \gets Q_i K_j^\top/\sqrt d$
            \State $m_i^{(j)} \gets \max(m_i^{(j-1)},\, \operatorname{rowmax}(S_{ij}))$
            \State $\widetilde P_{ij} \gets \exp(S_{ij} - m_i^{(j)})$
            \State $\ell_i^{(j)} \gets e^{m_i^{(j-1)} - m_i^{(j)}}\, \ell_i^{(j-1)} + \operatorname{rowsum}(\widetilde P_{ij})$
            \State $O_i \gets \operatorname{diag}(e^{m_i^{(j-1)} - m_i^{(j)}})\, O_i + \widetilde P_{ij}\, V_j$
        \Else
            \State $S_{ij} \gets \operatorname{Matmul}_{\mathrm{FP4}}(Q_i^q,K_j^q,S_{Q,i},S_{K,j})/\sqrt d$
            \State $m_i^{(j)} \gets \max(m_i^{(j-1)},\, \operatorname{rowmax}(S_{ij}))$
            \State $\widetilde P_{ij} \gets \exp(S_{ij} - m_i^{(j)})$
            \State $\ell_i^{(j)} \gets e^{m_i^{(j-1)} - m_i^{(j)}}\, \ell_i^{(j-1)} + \operatorname{rowsum}(\widetilde P_{ij})$
            \State $s^{(1)}_{P,ij} \gets \operatorname{rowmax}(\widetilde P_{ij})/(448 \times 6)$;\;
                   $(\widehat P_{ij}, S^{(2)}_{P,ij}) \gets \phi(\widetilde P_{ij}/s^{(1)}_{P,ij})$ // two-scale quantisation from SageAttention3
            \State $O_i \gets \operatorname{diag}(e^{m_i^{(j-1)} - m_i^{(j)}})\, O_i + s^{(1)}_{P,ij} \cdot \operatorname{Matmul}_{\mathrm{FP4}}(\widehat P_{ij}, V_j^q, S^{(2)}_{P,ij}, S_{V,j})$
        \EndIf
    \EndFor
    \State $O_i \gets \operatorname{diag}(\ell_i^{(T_k)})^{-1}\, O_i$
\EndFor
\State \Return $O = [O_1;\dots;O_{T_q}]$
\end{algorithmic}
\end{algorithm}

\subsection{Implementation and Optimisation on Hardware}
\noindent\textbf{Fused Mixed Precision Kernel.} The ThriftAttention mechanism is implemented as a single kernel. To reduce register pressure from supporting two precision paths in a single fused kernel, the FP16 query fragments are scoped only to the selected-block phase. The kernel first processes all non-selected KV blocks through the FP4 path using the FP4 query fragments. It then enters a separate FP16 helper routine for the promoted blocks where the FP16 query tile is loaded into registers. The same shared memory region is aliased for $K$, $V$ tiles across both precision paths, with double-buffered FP4 KV tiles to hide the latency of memory loads behind MMA instructions. Warps/CTAs whose KV range contains no top-$k$ blocks bypass the FP16 path entirely, avoiding both the HBM loads of FP16 $Q$, $K$, $V$ tiles and the associated register allocation. We implement our mixed precision kernel in CUDA C++.

\section{Experiments}
We validate the efficiency and performance of ThriftAttention across a diverse set of long-context evaluation tasks and model families. We benchmark kernel speed against FlashAttention-2 and SageAttention3, measure downstream accuracy at varying FP16 budgets on three long-context benchmarks, examine how recovery scales with context length on HELMET, and conduct a negative log-likelihood analysis on PG-19. We use block sizes $B_q = B_k = 64$ for all experiments.

\subsection{Efficiency and Effectiveness}
\begin{figure*}[!t]
  \centering
  \includegraphics[width=0.9\textwidth]{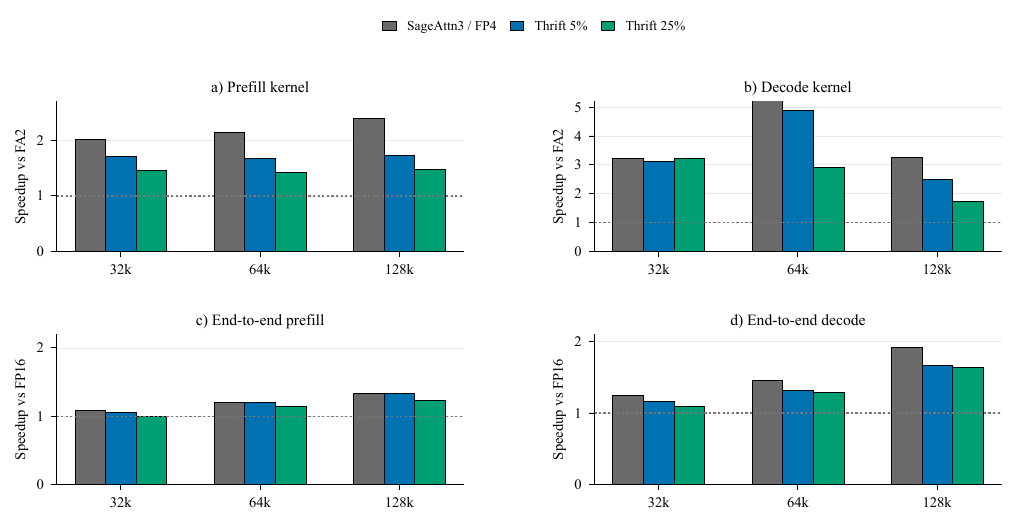}
  \caption{Kernel and end-to-end speedups over FlashAttention-2 for Prefill and Decode. $B=1$, $n_{heads}=32$, $D=128$. Qwen3-8B is used for end-to-end generation.}
  \label{fig:kernel-times}
\end{figure*}

Fig. \ref{fig:kernel-times} compares ThriftAttention's kernel and end-to-end generation latency to FlashAttention-2 and FP4 attention (SageAttention3) on an RTX PRO 6000. Note that FlashAttention-2 is the fastest attention variant supported on RTX 6000 Blackwell. For prefill, ThriftAttention achieves up to a $1.7\times$ kernel speedup over FlashAttention-2. These kernel gains translate to consistent end-to-end prefill improvements, reaching roughly $1.2\times$ speedup at 131k context lengths relative to FP16 attention. ThriftAttention's decode kernels achieve a $3\times$--$5.5\times$ speed-up compared to FlashAttention-2, with minimal overhead compared to full FP4 attention. This translates to a near $2\times$ end-to-end generation speedup over Qwen3-8B when using full FP16 for attention at 131k context length. By loading most blocks from the KV-cache in FP4, ThriftAttention reduces the dominant bottleneck in decoding.

\subsection{ThriftAttention Benchmark Evaluation}
\textbf{Experimental setup.} We evaluate ThriftAttention on three benchmarks: LongBench-v1~\citep{bai2024longbench}, HELMET~\citep{yen2024helmet}, and RULER~\citep{hsieh2024ruler}, across five models \citep{dubey2024llama3, yang2025qwen3, liu2026ministral3}. We compare performance to full FP4 attention (SageAttention3) and FP16.

\begin{table*}[t]
  \centering
  \small
  \footnotesize
  \setlength{\tabcolsep}{4pt}
  \begin{tabular}{ll cc cc cc}
  \toprule
  & & \multicolumn{2}{c}{RULER} & \multicolumn{2}{c}{LongBench v1} & \multicolumn{2}{c}{HELMET} \\
  \cmidrule(lr){3-4} \cmidrule(lr){5-6} \cmidrule(lr){7-8}
  Model & Method & Score & Recov. & Score & Recov. & Score & Recov. \\
  \midrule
  \multirow{5}{*}{Llama3.2 (3B)}
   & FP16        & 0.728 & 100  & 0.267 & 100  & 0.289 & 100  \\
   & FP4         & 0.321 & 0.0  & 0.132 & 0.0  & 0.101 & 0.0  \\
   & Top-$k$=5\% & 0.543 & 54.5 & 0.252 & 88.9 & 0.247    & 77.6  \\
   & Top-$k$=10\%& 0.548 & 55.8 & 0.249    & 86.5   & 0.251    & 79.6   \\
   & Top-$k$=25\%& 0.577 & 62.9 & 0.240 & 80.0 & 0.266    & 87.6  \\
  \midrule
  \multirow{5}{*}{Llama3.1 (8B)}
   & FP16        & 0.859 & 100  & 0.260 & 100   & 0.312 & 100  \\
   & FP4         & 0.385 & 0.0  & 0.133 & 0.0   & 0.132 & 0.0  \\
   & Top-$k$=5\% & 0.770 & 81.2 & 0.271 & 108.7 & 0.287 & 86.1 \\
   & Top-$k$=10\%& 0.778 & 82.9 & 0.275 & 111.8 & 0.286 & 85.6 \\
   & Top-$k$=25\%& 0.800 & 87.6 & 0.270 & 107.9 & 0.291 & 88.3 \\
  \midrule
  \multirow{5}{*}{Qwen3 (4B)}
   & FP16        & 0.795 & 100  & 0.260 & 100  & 0.249 & 100  \\
   & FP4         & 0.445 & 0.0  & 0.134 & 0.0  & 0.130 & 0.0  \\
   & Top-$k$=5\% & 0.749 & 86.9 & 0.254 & 95.2 & 0.232 & 85.7 \\
   & Top-$k$=10\%& 0.756 & 88.9 & 0.256 & 96.8 & 0.238 & 90.8 \\
   & Top-$k$=25\%& 0.768 & 92.3 & 0.257 & 97.6 & 0.246 & 97.5 \\
  \midrule
  \multirow{5}{*}{Qwen3 (8B)}
   & FP16        & 0.840 & 100  & 0.246 & 100  & 0.259 & 100   \\
   & FP4         & 0.484 & 0.0  & 0.128 & 0.0  & 0.127 & 0.0   \\
   & Top-$k$=5\% & 0.806 & 90.4 & 0.237 & 92.4 & 0.246 & 90.2  \\
   & Top-$k$=10\%& 0.815 & 93.0 & 0.241 & 95.8 & 0.268 & 106.8 \\
   & Top-$k$=25\%& 0.819 & 94.1 & 0.242 & 96.6 & 0.246 & 90.2  \\
  \midrule
  \multirow{5}{*}{Ministral 3 (8B)}
   & FP16        & 0.848 & 100   & 0.267 & 100  & 0.330 & 100   \\
   & FP4         & 0.548 & 0.0   & 0.130 & 0.0  & 0.159 & 0.0   \\
   & Top-$k$=5\% & 0.866 & 106.0 & 0.258 & 93.4 & 0.329 & 99.4  \\
   & Top-$k$=10\%& 0.864 & 105.3 & 0.262 & 96.4 & 0.331 & 100.6 \\
   & Top-$k$=25\%& 0.865 & 105.7 & 0.264 & 97.8 & 0.330 & 100.0 \\
  \bottomrule
  \end{tabular}
  \caption{Downstream accuracy of ThriftAttention at varying top-$k$ values across five models on long-context benchmarks. Recovery measures the fraction of the FP4--FP16 gap closed by each method.
}
  \label{tab:topk_longcontext}
\end{table*}

\textbf{Results.} 
Table \ref{tab:topk_longcontext} shows the efficacy of the approach across model families and evaluation benchmarks. Promoting only $5\%$ of blocks to FP16 results in an average recovery of $89.1\%$ of the FP4$\to$FP16 performance gap whilst $10\%$ and $25\%$ budgets push this further to $91.8\%$ and $92.4\%$ respectively. This indicates that the marginal quality returned per additional FP16 block yields diminishing returns once the most important blocks have been promoted to FP16. Results in Table \ref{tab:helmet_per_length} and Figure \ref{fig:nll-panel} provide further evidence of this. Benchmarks differ in their required FP16 budgets. LongBench v1 saturates at $5\%$ and barely moves with larger budgets. RULER scales steadily from $5\%$ to $25\%$. HELMET sits in between and scales unevenly.

\subsection{Sequence Length Experiments}
\textbf{Experimental Setup.}
We evaluate ThriftAttention on the HELMET benchmark across sequence lengths $L \in \{8192, \ldots, 131072\}$ and FP16 budgets $k/T_k \in \{5\%, 10\%, 25\% \}$.

\begin{table*}[t]
  \centering
  \small
  \footnotesize
  \setlength{\tabcolsep}{3pt}
  \begin{tabular}{ll cc cc cc cc cc}
  \toprule
  & & \multicolumn{2}{c}{8k} & \multicolumn{2}{c}{16k} & \multicolumn{2}{c}{32k} & \multicolumn{2}{c}{65k} & \multicolumn{2}{c}{131k} \\
  \cmidrule(lr){3-4} \cmidrule(lr){5-6} \cmidrule(lr){7-8} \cmidrule(lr){9-10} \cmidrule(lr){11-12}
  Model & Method & Score & Recov. & Score & Recov. & Score & Recov. & Score & Recov. & Score & Recov. \\
  \midrule
  \multirow{5}{*}{Llama3.2 (3B)}
   & FP16        & 0.295 & 100  & 0.312 & 100  & 0.279 & 100  & 0.285 & 100  & 0.274 & 100  \\
   & FP4         & 0.118 & 0.0  & 0.131 & 0.0  & 0.081 & 0.0  & 0.094 & 0.0  & 0.081 & 0.0    \\
   & Top-$k$=5\% & 0.276 & 89.3 & 0.279 & 81.8 & 0.247 & 83.8 & 0.228 & 70.2 &
  0.205 & 64.2 \\
   & Top-$k$=10\%& 0.278    & 90.2  & 0.282 & 83.4 & 0.245 & 82.8 & 0.233 & 72.8 & 0.215    & 69.4   \\
   & Top-$k$=25\%& 0.282    & 92.7   & 0.293 & 89.5 & 0.257 & 88.9 & 0.255 & 84.3 & 0.241    & 82.8   \\
  \midrule
  \multirow{5}{*}{Llama3.1 (8B)}
   & FP16        & 0.274 & 100  & 0.337 & 100   & 0.341 & 100  & 0.329 & 100  & 0.279 & 100  \\
   & FP4         & 0.138 & 0.0  & 0.161 & 0.0   & 0.149 & 0.0  & 0.127 & 0.0  & 0.088 & 0.0  \\
   & Top-$k$=5\% & 0.265 & 93.4 & 0.331 & 96.6  & 0.312 & 84.9 & 0.310 & 90.6 & 0.216 & 67.0 \\
   & Top-$k$=10\%& 0.261 & 90.4 & 0.334 & 98.3  & 0.318 & 88.0 & 0.303 & 87.1 & 0.211 & 64.4 \\
   & Top-$k$=25\%& 0.262 & 91.2 & 0.343 & 103.4 & 0.323 & 90.6 & 0.312 & 91.6 & 0.217 & 67.5 \\
  \midrule
  \multirow{5}{*}{Qwen3 (4B)}
   & FP16        & 0.321 & 100  & 0.270 & 100  & 0.280 & 100  & 0.222 & 100   & 0.154 & 100  \\
   & FP4         & 0.175 & 0.0  & 0.156 & 0.0  & 0.162 & 0.0  & 0.084 & 0.0   & 0.073 & 0.0  \\
   & Top-$k$=5\% & 0.303 & 87.7 & 0.252 & 84.2 & 0.279 & 99.2 & 0.195 & 80.4  & 0.130 & 70.4 \\
   & Top-$k$=10\%& 0.312 & 93.8 & 0.260 & 91.2 & 0.278 & 98.3 & 0.202 & 85.5  & 0.135 & 76.5 \\
   & Top-$k$=25\%& 0.318 & 97.9 & 0.268 & 98.2 & 0.275 & 95.8 & 0.222 & 100.0 & 0.148 & 92.6 \\
  \midrule
  \multirow{5}{*}{Qwen3 (8B)}
   & FP16        & 0.277 & 100   & 0.301 & 100   & 0.272 & 100   & 0.232 & 100  & 0.214 & 100  \\
   & FP4         & 0.143 & 0.0   & 0.139 & 0.0   & 0.154 & 0.0   & 0.103 & 0.0  & 0.096 & 0.0  \\
   & Top-$k$=5\% & 0.275 & 98.5  & 0.294 & 95.7  & 0.271 & 99.2  & 0.198 & 73.6 & 0.192 & 81.4 \\
   & Top-$k$=10\%& 0.299 & 116.4 & 0.320 & 111.7 & 0.296 & 120.3 & 0.216 & 87.6 & 0.209 & 95.8 \\
   & Top-$k$=25\%& 0.264 & 90.3  & 0.298 & 98.1  & 0.241 & 73.7  & 0.217 & 88.4 & 0.210 & 96.6 \\
  \midrule
  \multirow{5}{*}{Ministral 3 (8B)}
   & FP16        & 0.325 & 100   & 0.318 & 100   & 0.343 & 100   & 0.338 & 100  & 0.326 & 100   \\
   & FP4         & 0.141 & 0.0   & 0.155 & 0.0   & 0.169 & 0.0   & 0.174 & 0.0  & 0.157 & 0.0   \\
   & Top-$k$=5\% & 0.323 & 98.9  & 0.318 & 100.0 & 0.347 & 102.3 & 0.327 & 93.3 & 0.331 & 103.0 \\
   & Top-$k$=10\%& 0.328 & 101.6 & 0.318 & 100.0 & 0.337 & 96.6  & 0.334 & 97.6 & 0.338 & 107.1 \\
   & Top-$k$=25\%& 0.323 & 98.9  & 0.318 & 100.0 & 0.344 & 100.6 & 0.331 & 95.7 & 0.334 & 104.7 \\
  \bottomrule
  \end{tabular}
  \caption{Per-length HELMET accuracy of ThriftAttention at varying top-$k$ values across five models.}
  \label{tab:helmet_per_length}
\end{table*}

\textbf{Per-length recovery.} The sequence-level ablation in Table \ref{tab:helmet_per_length} demonstrates that FP4 quality degrades relative to FP16 performance as sequence length increases. Llama3.1-8B drops from $50\%$ FP4 retention at 8k to $32\%$ at 131k and the two Qwen variants and Llama3.2-3B show comparable falls. Relative to FP4, ThriftAttention's performance improves at longer contexts, increasing from $2.00\times$ FP4 performance at 8k to $2.2\times$ at 131k with a $5\%$ FP16 block budget. This increase in relative improvement over FP4 demonstrates that the approach becomes increasingly valuable as context length grows. Table~\ref{tab:topk_performance} in Appendix~\ref{app:short_context} shows that FP4 is substantially closer to FP16 on short-context evaluation benchmarks, further supporting ThriftAttention's focus on long-context settings. We analyse context length scaling trends at finer granularity in the following section.

\subsection{Negative Log-Likelihood Analysis}
We conduct negative log-likelihood (NLL) analysis to thoroughly examine the context-length scaling trend in ThriftAttention. We record per-token NLL on 300 documents sampled from PG-19~\citep{rae2020compressive} across context lengths and FP16 budgets.

\begin{figure*}[t]
  \centering
  \includegraphics[width=0.6\textwidth]{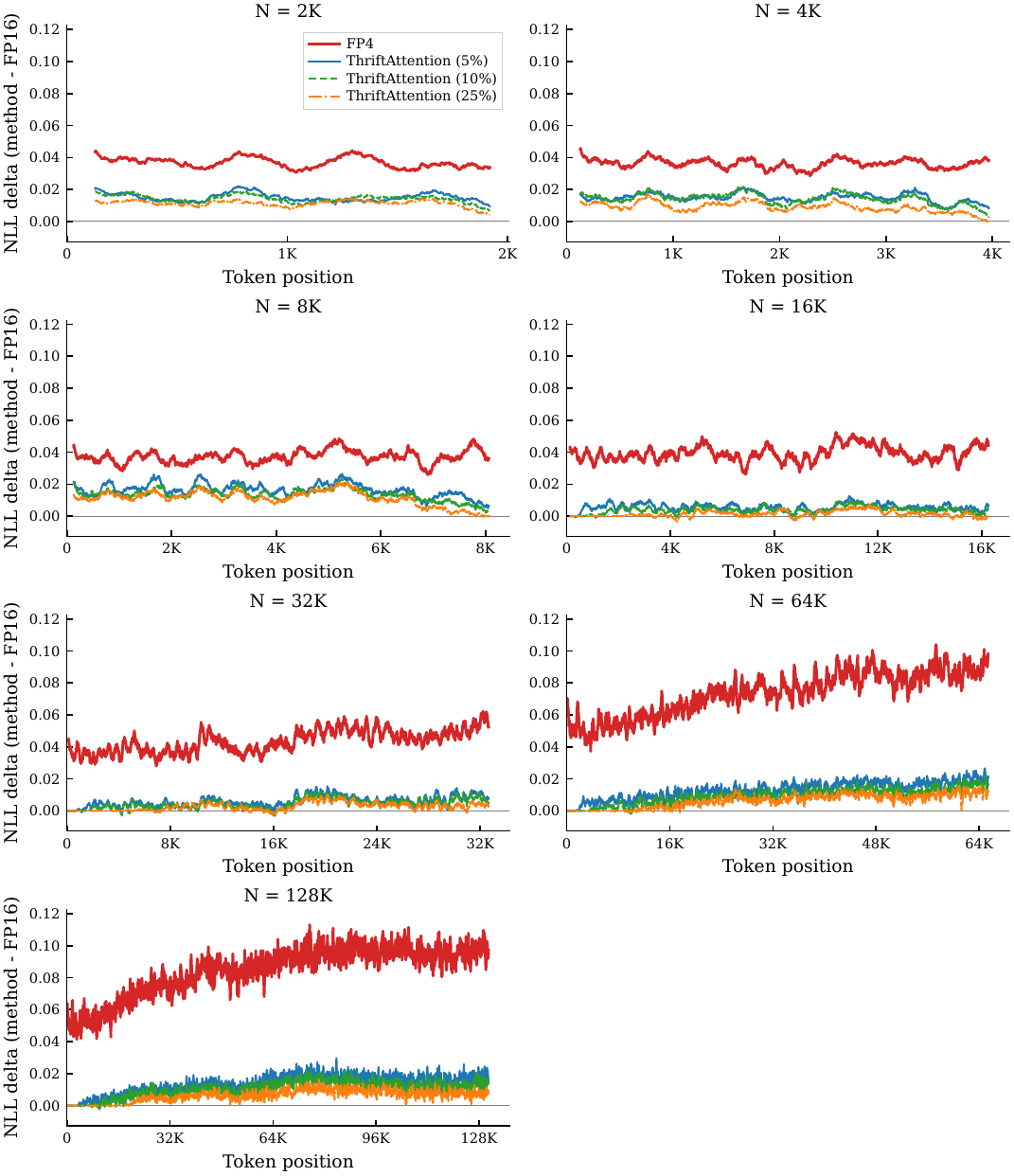}
  \caption{Per-token negative log-likelihood increase over the FP16 baseline ($\Delta\text{NLL} = \text{NLL}_{\text{method}} - \text{NLL}_{\text{FP16}}$), over PG-19 documents across context lengths for Qwen3-8B. }
  \label{fig:nll-panel}
\end{figure*}

\paragraph{ThriftAttention more effective at long contexts.} Figure~\ref{fig:nll-panel} reveals a systematic degradation in FP4 quality as context length increases. For sequence lengths below $16$k, the FP4 $\Delta\text{NLL}$ remains approximately constant at $0.04$ across token positions. Beyond $32$k a positional dependence appears, and at $64$k and $128$k the FP4 $\Delta\text{NLL}$ degrades to $0.10$ for tokens at the end of the sequence. ThriftAttention mitigates this effect, reducing $\Delta\text{NLL}$ to  $\leq 0.02$ across context lengths. Consequently the relative reduction in $\Delta\text{NLL}$ provided by ThriftAttention grows from approximately $2\times$ at $8$k to approximately $5\times$ at the end of a $128$k sequence.

The widening advantage at longer contexts can be attributed to the increasingly concentrated structure of attention itself. As sequence length grows, the total number of query-key block interactions increases quadratically, while the subset of interactions carrying most of the attention mass grows more slowly. This means at longer contexts the most sensitive query-key interactions occupy a smaller fraction of all possible block interactions. A fixed fractional top-$k$ budget therefore captures an increasingly large share of the attention mass at longer contexts.

\subsection{Comparison to Sparse Attention Baselines at Matched Compute}
We evaluate ThriftAttention against inference-time sparse-attention approaches to directly compare our mixed-precision technique to sparsity mechanisms. We compare ThriftAttention at $5\%$ to Quest~\citep{tang2024quest} and sparse top-$k$, both running at a sparsity ratio of $71.3\%$. This yields an equivalent total FLOP budget between sparse approaches and ThriftAttention at a $5\%$ FP16 budget.

\begin{table}[t]
      \centering
      \small
      \begin{tabular}{lcccc}
        \toprule
        Method & FP16 \% & FP4 \% & Skipped \% & Score \\
        \midrule
        ThriftAttention 5\% & 5.0 & 95.0 & 0.0 & 0.599  \\
        Sparse Top-$k$ 28.7\% & 28.7 & 0.0 & 71.3 & 0.036  \\
        Quest 28.7\% & 28.7 & 0.0 & 71.3 & 0.142  \\
        \bottomrule
      \end{tabular}
      \caption{Average performance at sequence length 65,536 on a subset of HELMET tasks under matched FP16-equivalent compute. ThriftAttention computes all blocks using mixed precision, whereas sparse methods compute only 28.7\% of blocks and skip the remainder.}
      \label{tab:avg_helmet_65536}
  \end{table}

Table \ref{tab:avg_helmet_65536} shows that ThriftAttention outperforms inference-time sparsity approaches at equivalent inference efficiency. This supports a core claim of ThriftAttention. Sparse methods can incur large tail errors because missed blocks are removed entirely, whereas ThriftAttention degrades more smoothly by retaining all interactions in low precision. In ThriftAttention, the same missed block remains available in FP4, so its contribution is degraded by quantisation rather than deleted. Thus, at matched compute, preserving the full attention support in FP4 is more effective than sparsifying aggressively and computing only a small subset in FP16. 

\section{Limitations and Future Work} 
Our current kernel implementation targets consumer Blackwell GPUs. Extending ThriftAttention to data-center Blackwell would allow the approach to exploit SM100 features such as increased asynchrony which may move FP4 performance closer to the $4\times$ theoretical throughput advantage over FP16. Whilst our kernel improves inference latency, it increases KV-cache memory footprint by $28\%$ by storing FP16 and FP4 caches. ThriftAttention is currently designed for inference acceleration only. Existing 4-bit training methods~\citep{nvidia2026nvfp4training, chmiel2025fp4alltheway, castro2025quartet} typically retain attention in higher precision. Selectively promoting sensitive interactions in the forward and backward attention computation to FP16 could help address stability issues in sub-byte attention training.

\section{Conclusion}
We introduced \textbf{ThriftAttention}, a selective mixed-precision attention mechanism for long-context FP4 inference. By selectively promoting only a small number of blocks in the attention computation to FP16, we prevent systematic performance degradation in long-context settings. We demonstrate this behaviour across model families and evaluation benchmarks. Overall, selective precision offers a practical path toward long-context inference that approaches FP4 latency while preserving near-FP16 quality.

\newpage
\bibliographystyle{plainnat}
\bibliography{references}
\newpage
\appendix

\section{Heuristic Ablation}
We conducted a heuristic ablation of the ThriftAttention heuristic against randomly promoted blocks and diagonal-only selection. 

\begin{table}[h]
    \centering
    \small
    \begin{tabular}{lcc}
      \toprule
      Method & Precision mix & Score \\
      \midrule
      ThriftAttention & 5\% FP16 / 95\% FP4 & 0.599  \\
      Random block selection & 5\% FP16 / 95\% FP4 & 0.407  \\
      Diagonal block selection & 5\% FP16 / 95\% FP4 & 0.521 \\
      \bottomrule
    \end{tabular}
    \caption{Heuristic ablation on a subset of HELMET tasks at sequence length 65,536 with a $5\%$ FP16 budget. ThriftAttention outperforms random and diagonal block selection. The HELMET tasks evaluated are \texttt{json\_kv}, \texttt{kilt\_popqa\_3}, and \texttt{long\_narrative\_qa}.}
    \label{tab:avg_helmet_heuristic}
\end{table}

\section{Experiment Design}
\label{app:experiment_design}

\paragraph{Code and environment.}
Experiments use CUDA 12.8, PyTorch 2.8.0, and a single NVIDIA RTX PRO 6000
Blackwell GPU with 96GB GPU memory. The full set of reported downstream
benchmark evaluations required approximately 600 GPU-hours. The NLL analysis
required approximately 5 GPU-hours. We did not use a larger internal cluster or
additional undisclosed compute for the reported results.

\paragraph{Models.}
\begin{center}
\small
\begin{tabular}{ll}
\toprule
Paper label & Checkpoint ID \\
\midrule
Llama3.2 (3B) & \texttt{meta-llama/Llama-3.2-3B} \\
Llama3.1 (8B) & \texttt{meta-llama/Llama-3.1-8B} \\
Qwen3 (4B) & \texttt{Qwen/Qwen3-4B} \\
Qwen3 (8B) & \texttt{Qwen/Qwen3-8B} \\
Ministral 3 (8B) & \texttt{mistralai/Ministral-3-8B-Base-2512} \\
\bottomrule
\end{tabular}
\end{center}

\paragraph{Context, and decoding.}
The maximum context is 131,072 tokens, clamped by the model limit.  Qwen3 uses base RoPE up to 32,768 tokens and YaRN beyond that whilst Llama and Ministral use checkpoint RoPE settings. All generation uses greedy argmax decoding.

\paragraph{ThriftAttention settings.}
All experiments use block sizes $B_q=B_k=64$.  Target FP16 budgets are $5\%$, $10\%$, and $25\%$.  For a sequence with $n=\lfloor L/64\rfloor$ key blocks, the integer top-$k$ is chosen by
\[
  (k n-k(k-1)/2)/(n(n+1)/2) \approx f,
\]
rounded and clamped to $[1,n]$.

\paragraph{Benchmarks.}
LongBench v1 is evaluated through lm-eval-harness on the English subset:
\texttt{narrativeqa}, \texttt{qasper}, \texttt{multifieldqa\_en},
\texttt{hotpotqa}, \texttt{2wikimqa}, \texttt{musique},
\texttt{gov\_report}, \texttt{qmsum}, \texttt{multi\_news}, \texttt{trec},
\texttt{triviaqa}, \texttt{samsum}, \texttt{passage\_count},
\texttt{passage\_retrieval\_en}, \texttt{lcc}, and \texttt{repobench-p};
the reported score is \texttt{\_overall.weighted\_avg}.  RULER uses the local NVIDIA/RULER port with 100 samples per task-length cell over 13 tasks: 8 NIAH variants, \texttt{vt}, \texttt{cwe}, \texttt{fwe}, \texttt{qa\_1}, and \texttt{qa\_2}, at lengths
\(\{4096,8192,16384,32768,65536,131072\}\).  HELMET uses the local HELMET
config snapshot with 50 samples per task-length cell over recall
(\texttt{json\_kv}), RAG (\texttt{kilt\_nq}, \texttt{kilt\_triviaqa},
\texttt{kilt\_hotpotqa}, \texttt{kilt\_popqa\_3}), reranking
(\texttt{msmarco\_rerank\_psg}), LongQA (\texttt{narrativeqa},
\texttt{infbench\_qa}, \texttt{infbench\_choice}), summarization
(\texttt{infbench\_sum}, \texttt{multi\_lexsum}), and five ICL tasks, at
\(\{8192,16384,32768,65536,131072\}\).

\paragraph{NLL and sequence-length analysis.}
NLL uses 300 packed sequences from \texttt{emozilla/pg19}, seed 42, with EOS inserted between documents.  We evaluate Qwen3-8B at
\(\{2048,4096,8192,16384,32768,65536,131072\}\) using teacher-forced next-token cross-entropy; no generation is used. The sequence-length analysis uses the same HELMET settings and greedy decoding as above.

\section{Short-context benchmark results}
\label{app:short_context}

We also evaluate on short-context reasoning and knowledge benchmarks:
BBH~\citep{suzgun2023challenging}, MMLU-Pro~\citep{wang2024mmlupro}, 
and GSM8K~\citep{cobbe2021gsm8k}.
\begin{table*}[h]
  \centering
  \small
  \footnotesize
  \setlength{\tabcolsep}{3.5pt}
  \begin{tabular}{ll ccc ccc ccc}
  \toprule
  & & \multicolumn{3}{c}{BBH$_{(\text{EM})}$} & \multicolumn{3}{c}{MMLU-Pro$_{(\text{EM})}$} & \multicolumn{3}{c}{GSM8K-COT$_{(\text{EM})}$} \\
  \cmidrule(lr){3-5} \cmidrule(lr){6-8} \cmidrule(lr){9-11}
  Model & Method & Score & Recov. & FP16\% & Score & Recov. & FP16\% & Score & Recov. & FP16\% \\
  \midrule
  \multirow{5}{*}{Llama3.2 (3B)}
   & FP16      & 0.470 & --   & 100  & 0.248 & --   & 100 & 0.279 & --    & 100 \\
   & FP4       & 0.408 & 0.0  & 0    & 0.207 & 0.0  & 0   & 0.178 & 0.0   & 0 \\
   & Top-$k$=1 & 0.465 & 91.9 & 10.8 & 0.233 & 63.4 & 8.2 & \textbf{0.282} & 103.0 & 6.3  \\
   & Top-$k$=2 & 0.462 & 87.1 & 21.3 & \textbf{0.242} & 85.4 & 16.0 & 0.273 & 94.0 & 12.4 \\
   & Top-$k$=4 & \textbf{0.465} & 91.9 & 40.1 & 0.242 & 85.4 & 30.5 & 0.264 & 85.0 & 24.6 \\
  \midrule
  \multirow{5}{*}{Llama3.1 (8B)}
   & FP16      & 0.630 & --   & 100  & 0.359 & --   & 100 & 0.547 & --   & 100 \\
   & FP4       & 0.516 & 0.0  & 0    & 0.286 & 0.0  & 0   & 0.417 & 0.0  & 0 \\
   & Top-$k$=1 & 0.614 & 86.0 & 11.1 & 0.336 & 68.5 & 7.2 & 0.527 & 84.8 & 5.2 \\
   & Top-$k$=2 & 0.619 & 90.4 & 21.6 & 0.345 & 80.8 & 17.0 & 0.520 & 79.5 & 10.1 \\
   & Top-$k$=4 & \textbf{0.624} & 94.7 & 40.6 & \textbf{0.346} & 82.2 & 31.6 & \textbf{0.541} & 95.9 & 22.9 \\
  \midrule
  \multirow{5}{*}{Qwen3 (4B)}
   & FP16      & 0.752 & --   & 100  & 0.546 & --   & 100 & 0.880 & --   & 100 \\
   & FP4       & 0.675 & 0.0  & 0    & 0.509 & 0.0  & 0   & 0.835 & 0.0  & 0 \\
   & Top-$k$=1 & 0.731 & 72.7 & 11.5 & 0.530 & 56.8 & 10.2 & 0.854 & 43.3 & 5.4 \\
   & Top-$k$=2 & \textbf{0.732} & 74.0 & 21.9 & 0.536 & 73.0 & 18.0 & \textbf{0.863} & 61.7 & 11.0 \\
   & Top-$k$=4 & 0.726 & 66.2 & 42.0 & \textbf{0.537} & 75.7 & 33.5 & 0.861 & 56.7 & 21.7 \\
  \midrule
  \multirow{5}{*}{Qwen3 (8B)}
   & FP16      & 0.795 & --   & 100  & 0.597 & --   & 100 & 0.901 & --   & 100 \\
   & FP4       & 0.738 & 0.0  & 0    & 0.564 & 0.0  & 0   & 0.868 & 0.0  & 0 \\
   & Top-$k$=1 & 0.775 & 64.9 & 10.4 & 0.581 & 51.5 & 7.9 & 0.873 & 13.9 & 5.1 \\
   & Top-$k$=2 & \textbf{0.782} & 77.2 & 20.2 & 0.587 & 69.7 & 16.7 & 0.883 & 46.5 & 11.3 \\
   & Top-$k$=4 & 0.772 & 59.6 & 40.9 & \textbf{0.588} & 72.7 & 30.7 & \textbf{0.896} & 86.0 & 22.4 \\
  \bottomrule
  \end{tabular}
  \caption{Downstream accuracy of ThriftAttention at varying top-$k$ values across four models evaluated on BBH, MMLU-Pro, and GSM8K. Recov.\ denotes the percentage recovery from FP4 to FP16 performance.}
  \label{tab:topk_performance}
\end{table*}
\end{document}